\documentclass{llncs}
\usepackage{amsmath, amsfonts, graphicx, booktabs, tabularx, multirow, siunitx, xcolor}
\usepackage{hyperref}

\graphicspath{{./figures/}}
\begin{document}

\title{3D Segmentation with Exponential Logarithmic Loss for Highly Unbalanced Object Sizes\thanks{This paper was accepted by the International Conference on Medical Image Computing and Computer-Assisted Intervention -- MICCAI 2018 (oral presentation). The final publication is available at Springer via \url{https://doi.org/10.1007/978-3-030-00931-1_70}}}
\author{Ken C. L. Wong, Mehdi Moradi\thanks{Corresponding author.}, Hui Tang, and Tanveer Syeda-Mahmood}
\institute{IBM Research -- Almaden Research Center, San Jose, CA, USA\\
\email{mmoradi@us.ibm.com}
}

\maketitle              

\begin{abstract}
With the introduction of fully convolutional neural networks, deep learning has raised the benchmark for medical image segmentation on both speed and accuracy, and different networks have been proposed for 2D and 3D segmentation with promising results. Nevertheless, most networks only handle relatively small numbers of labels ($<$10), and there are very limited works on handling highly unbalanced object sizes especially in 3D segmentation. In this paper, we propose a network architecture and the corresponding loss function which improve segmentation of very small structures. By combining skip connections and deep supervision with respect to the computational feasibility of 3D segmentation, we propose a fast converging and computationally efficient network architecture for accurate segmentation. Furthermore, inspired by the concept of focal loss, we propose an exponential logarithmic loss which balances the labels not only by their relative sizes but also by their segmentation difficulties. We achieve an average Dice coefficient of 82\% on brain segmentation with 20 labels, with the ratio of the smallest to largest object sizes as 0.14\%. Less than 100 epochs are required to reach such accuracy, and segmenting a 128$\times$128$\times$128 volume only takes around 0.4 s.
\end{abstract}
\section{Introduction}

With the introduction of fully convolutional neural networks (CNNs), deep learning has raised the benchmark for medical image segmentation on both speed and accuracy \cite{Conference:Ronneberger:MICCAI2015}. Different 2D \cite{Conference:Mehta:ISBI2017,Conference:Roy:MICCAI2017} and 3D \cite{Conference:Cicek:MICCAI2016,Conference:Milletari:3DV2016,Journal:Dou:MedIA2017,Conference:Tang:SPIE2018} networks were proposed to segment various anatomies such as the heart, brain, liver, and prostate from medical images. Regardless of the promising results of these networks, 3D CNN image segmentation is still challenging. Most networks were applied on datasets with small numbers of labels ($<$10) especially in 3D segmentation. When more detailed segmentation is required with much more anatomical structures, previously unseen issues, such as computational feasibility and highly unbalanced object sizes, need to be addressed by new network architectures and algorithms.

There are only a few frameworks proposed for highly unbalanced labels. In \cite{Conference:Roy:MICCAI2017}, a 2D network architecture was proposed to segment all slices of a 3D brain volume. Error corrective boosting was introduced to compute label weights that emphasize parameter updates on classes with lower validation accuracy. Although the results were promising, the label weights were only applied to the weighted cross-entropy but not the Dice loss, and the stacking of 2D results for 3D segmentation may result in inconsistency among consecutive slices.

In \cite{Conference:Sudre:DLMIA2017}, the generalized Dice loss was used as the loss function. Instead of computing the Dice loss of each label, the weighted sum of the products over the weighted sum of the sums between the ground-truth and predicted probabilities was computed for the generalized Dice loss, with the weights inversely proportional to the label frequencies. In fact, the Dice coefficient is unfavorable to small structures as a few pixels of misclassification can lead to a large decrease of the coefficient, and this sensitivity is irrelevant to the relative sizes among structures. Therefore, balancing by label frequencies is nonoptimal for Dice losses.

To address the issues of highly unbalanced object sizes and computational efficiency in 3D segmentation, we have two key contributions in this paper. \textbf{I)} We propose the exponential logarithmic loss function. In \cite{Journal:Lin:arXiv2017}, to handle the highly unbalanced dataset of a two-class image classification problem, a modulating factor computed solely from the softmax probability of the network output is multiplied by the weighted cross-entropy to focus on the less accurate class. Inspired by this concept of balancing classification difficulties, we propose a loss function comprising the logarithmic Dice loss which intrinsically focuses more on less accurately segmented structures. The nonlinearities of the logarithmic Dice loss and the weighted cross-entropy can be further controlled by the proposed exponential parameters. In this manner, the network can achieve accurate segmentation on both small and large structures. \textbf{II)} We propose a fast converging and computationally efficient network architecture by combining the advantages of skip connections and deep supervision, which has only about 1/14 of the parameters of, and is twice as fast as, the V-Net  \cite{Conference:Milletari:3DV2016}. Experiments were performed on brain magnetic resonance (MR) images with 20 highly unbalanced labels. Combining these two innovations achieved an average Dice coefficient of 82\% with the average segmentation time as 0.4 s.

\section{Methodology}

\subsection{Proposed Network Architecture}

\begin{figure}[t]
    \centering
    \begin{minipage}[b]{1\linewidth}
      \centering
      \includegraphics[width=1\linewidth]{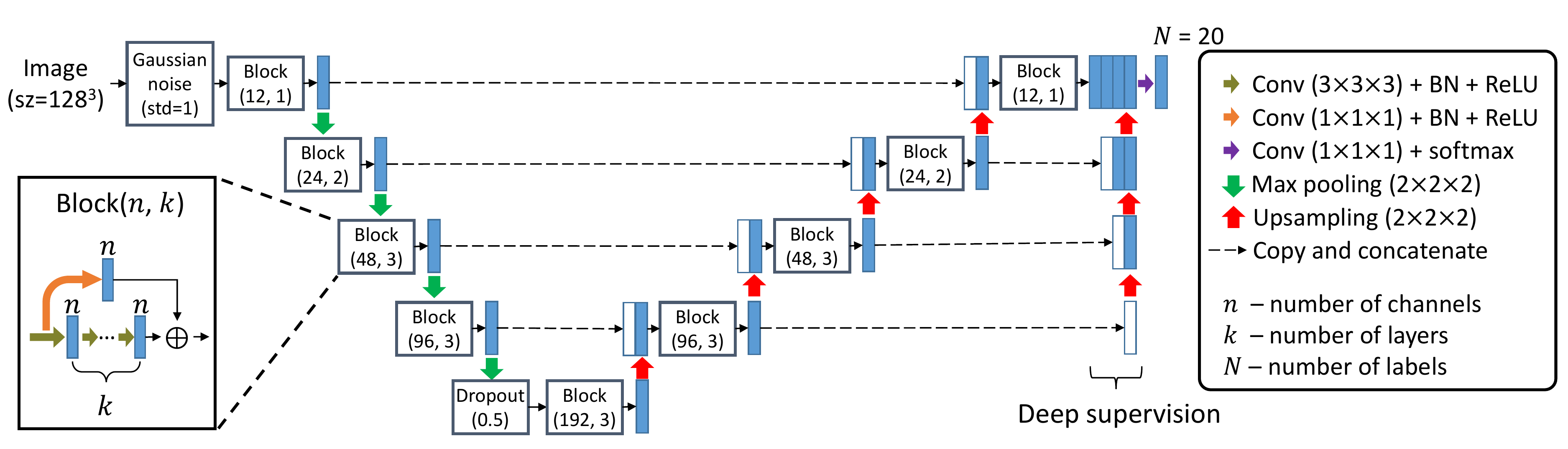}
    \end{minipage}
    \caption{Proposed network architecture optimized for 3D segmentation. Blue and white boxes indicate operation outputs and copied data, respectively.}
    \label{fig:network}
\end{figure}

3D segmentation networks require much more computational resources than 2D networks. Therefore, we propose a network architecture which aims at accurate segmentation and fast convergence with respect to limited resources (Fig. \ref{fig:network}). Similar to most segmentation networks, our network comprises the encoding and decoding paths. The network is composed of convolutional blocks, each comprises $k$ cascading 3$\times$3$\times$3 convolutional layers of $n$ channels associated with batch normalization (BN) and rectified linear units (ReLU). For better convergence, a skip connection with a 1$\times$1$\times$1 convolutional layer is used in each block. Instead of concatenation, we add the two branches together for less memory consumption, so the block allows efficient multi-scale processing and deeper networks can be trained. The number of channels ($n$) is doubled after each max pooling and is halved after each upsampling. More layers ($k$) are used with tensors of smaller sizes so that more abstract knowledge can be learned with feasible memory use. Feature channels from the encoding path are concatenated with the corresponding tensors in the decoding path for better convergence. We also include a Gaussian noise layer and a dropout layer to avoid overfitting.

Similar to \cite{Conference:Mehta:ISBI2017}, we utilize deep supervision which allows more direct backpropagation to the hidden layers for faster convergence and better accuracy \cite{Conference:Lee:AISTATS2015}. Although deep supervision significantly improves convergence, it is memory expensive especially in 3D networks. Therefore, we omit the tensor from the block with the most channels (Block(192, 3)) so that training can be performed on a GPU with 12 GB of memory. A final layer of 1$\times$1$\times$1 convolution with the softmax function provides the segmentation probabilities.

\subsection{Exponential Logarithmic Loss}
\label{sec:loss}

We propose a loss function which improves segmentation on small structures:
\begin{gather}
\label{eq:weighted_loss}
    L_\mathrm{Exp} = w_{\mathrm{Dice}} L_\mathrm{Dice} + w_\mathrm{Cross} L_\mathrm{Cross}
\end{gather}
with $w_{\mathrm{Dice}}$ and $w_\mathrm{Cross}$ the respective weights of the exponential logarithmic Dice loss ($L_\mathrm{Dice}$) and the weighted exponential cross-entropy ($L_\mathrm{Cross}$):
\begin{gather}
\label{eq:dice_loss}
    L_\mathrm{Dice} = \mathbf{E}\left[\left( - \ln
    (\mathrm{Dice}_i)
    \right)^{\gamma_\mathrm{Dice}}\right]
    \ \textrm{with}\ \
    \mathrm{Dice}_i = \frac{2 \left(\sum\nolimits_\mathbf{x} \delta_{il}(\mathbf{x})\ p_i(\mathbf{x})\right) + \epsilon}
    {\left(\sum\nolimits_\mathbf{x} \delta_{il}(\mathbf{x}) + p_i(\mathbf{x})\right) + \epsilon}  \\
\label{eq:entropy_loss}
    L_\mathrm{Cross} = \mathbf{E}\left[ w_l \left(- \ln(p_l(\mathbf{x}))\right)^{\gamma_\mathrm{Cross}} \right]
\end{gather}
with $\mathbf{x}$ the pixel position and $i$ the label. $l$ is the ground-truth label at $\mathbf{x}$. $\mathbf{E}[\bullet]$ is the mean value with respect to $i$ and $\mathbf{x}$ in $L_\mathrm{Dice}$ and $L_\mathrm{Cross}$, respectively. $\delta_{il}(\mathbf{x})$ is the Kronecker delta which is 1 when $i=l$ and 0 otherwise. $p_{i}(\mathbf{x})$ is the softmax probability which acts as the portion of pixel $\mathbf{x}$ owned by label $i$ when computing $\mathrm{Dice}_i$. $\epsilon = 1$ is the pseudocount for additive smoothing to handle missing labels in training samples. $w_{l} = \left((\sum\nolimits_k f_k)/{f_l}\right)^{0.5}$, with $f_k$ the frequency of label $k$, is the label weight for reducing the influences of more frequently seen labels. $\gamma_\mathrm{Dice}$ and $\gamma_\mathrm{Cross}$ further control the nonlinearities of the loss functions, and we use $\gamma_\mathrm{Dice} = \gamma_\mathrm{Cross} = \gamma$ here for simplicity.

\begin{figure}[t]
    \footnotesize
    \centering
    \begin{minipage}[b]{0.35\linewidth}
      \centering
      \includegraphics[width=1\linewidth]{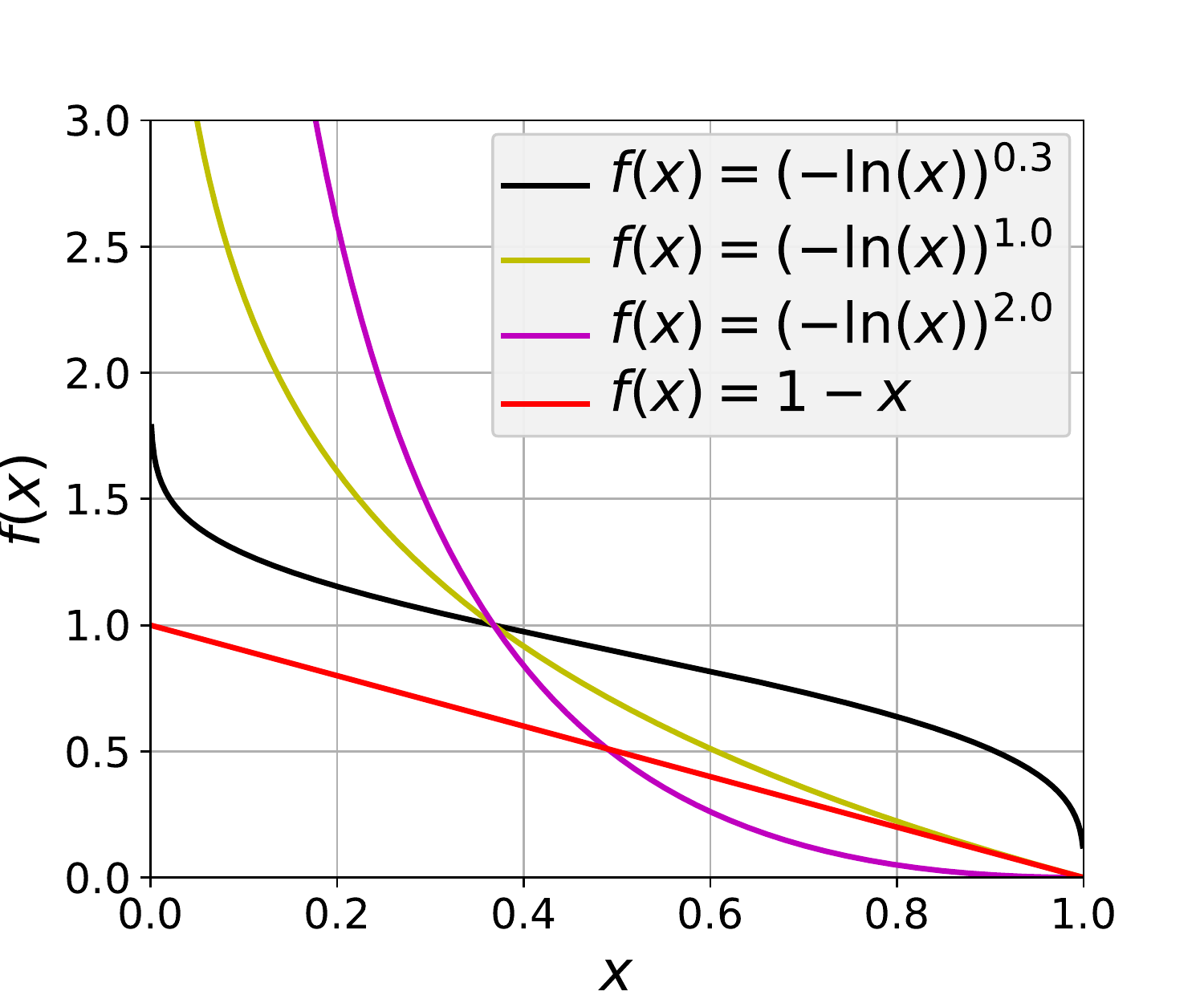} \\
    \end{minipage}
    \caption{Loss functions with different nonlinearities, where $x$ can be $\mathrm{Dice}_i$ or $p_l(\mathbf{x})$.}
    \label{fig:plot_loss}
\end{figure}

The use of the Dice loss in CNN was proposed in \cite{Conference:Milletari:3DV2016}. The Dice coefficient is unfavorable to small structures as misclassifying a few pixels can lead to a large decrease of the coefficient. The use of label weights cannot alleviate such sensitivity as it is irrelevant to the relative object sizes, and the Dice coefficient is already a normalized metric. Therefore, instead of size differences, we use the logarithmic Dice loss which focuses more on less accurate labels. Fig. \ref{fig:plot_loss} shows a comparison between the linear ($\mathbf{E}\left[1 - \mathrm{Dice}_i\right]$) and logarithmic Dice loss.


We provide further control on the nonlinearities of the losses by introducing the exponents $\gamma_\mathrm{Dice}$ and $\gamma_\mathrm{Cross}$. In \cite{Journal:Lin:arXiv2017}, a modulating factor, $(1 - p_l)^\gamma$, is multiplied by the weighted cross-entropy to become $w_l(1 - p_l)^\gamma(-\ln(p_l))$ for two-class image classification. Apart from balancing the label frequencies using the label weights $w_l$, this focal loss also balances between easy and hard samples. Our exponential loss achieves a similar goal. With $\gamma > 1$, the loss focuses more on less accurate labels than the logarithmic loss (Fig. \ref{fig:plot_loss}). Although the focal loss works well for the two-class image classification in \cite{Journal:Lin:arXiv2017}, we got worse results when applying to our segmentation problem with 20 labels. This may be caused by the over suppression of the loss function when the label accuracy becomes high. In contrast, we could get better results with $0 < \gamma < 1$. Fig. \ref{fig:plot_loss} shows that when $\gamma = 0.3$, there is an inflection point around $x = 0.5$, where $x$ can be $\mathrm{Dice}_i$ or $p_l(\mathbf{x})$. For $x < 0.5$, this loss behaves similarly to the losses with $\gamma \geq 1$ with decreasing gradient magnitude as $x$ increases. This trend reverses for $x > 0.5$ with increasing gradient magnitude. In consequence, this loss encourages improvements at both low and high prediction accuracy. This characteristics is the reason of using the proposed exponential form instead of the one in \cite{Journal:Lin:arXiv2017}.

\subsection{Training Strategy}

Image augmentation is used to learn invariant features and avoid overfitting. As realistic nonrigid deformation is difficult to implement and computationally expensive, we limit the augmentation to rigid transformations including rotation (axial, $\pm$\ang{30}), shifting ($\pm$20\%), and scaling ([0.8, 1.2]). Each image has an 80\% chance to be transformed in training, thus the number of augmented images is proportional to the number of epochs. The optimizer Adam is used with the Nesterov momentum for fast convergence, with the learning rate as 10$^{-3}$, batch size as one, and 100 epochs. A TITAN X GPU with 12 GB of memory is used.

\section{Experiments}

\subsection{Data and Experimental Setups}

A dataset of 43 3D brain MR images from different patients was neuroanatomically labeled to provide the training and validation samples. The images were produced by the T1-weighted MP-RAGE pulse sequence which provides high tissue contrast. They were manually segmented by highly trained experts with the results reviewed by a consulting neuroanatomist. Each segmentation had 19 semantic labels of brain structures, thus 20 labels with the background included (Table \ref{table:brain}(a)). As there were various image sizes (128 to 337) and spacings (0.9 to 1.5 mm), each image was resampled to isotropic spacing using the minimum spacing, zero padded on the shorter sides, and resized to 128$\times$128$\times$128.

Table \ref{table:brain}(a) shows that the labels were highly unbalanced. The background occupied 93.5\% of an image on average. Without the background, the relative sizes of the smallest and largest structures were 0.07\% and 50.24\%, respectively, thus a ratio of 0.14\%.

We studied six loss functions using the proposed network, and applied the best one to the V-Net architecture \cite{Conference:Milletari:3DV2016}, thus a total of seven cases were studied. For $L_\mathrm{Exp}$, we set $w_{\mathrm{Dice}} = 0.8$ and $w_{\mathrm{Cross}} = 0.2$ as they provided the best results. Five sets of data were generated by shuffling and splitting the dataset, with 70\% for training and 30\% for validation in each set. Experiments were performed on all five sets of data for each case studied for more statistically sound results. The actual Dice coefficients, not the $\mathrm{Dice}_i$ in (\ref{eq:dice_loss}), were computed for each validation image. Identical setup and training strategy were used in all experiments.

\begin{table*}[t]
\caption{Semantic brain segmentation. (a) Semantic labels and their relative sizes on average (\%) without the background. CVL represents cerebellar vermal lobules. The background occupied 93.5\% of an image on average. (b) Dice coefficients between prediction and ground truth averaged from five experiments (format: mean$\pm$std\%). The best results are highlighted in blue. $w_{\mathrm{Dice}} = 0.8$ and $w_{\mathrm{Cross}} = 0.2$ for all $L_\mathrm{Exp}$.}
\label{table:brain}

\smallskip
\fontsize{6}{7}\selectfont
\centering

\newcolumntype{R}{>{\raggedleft\arraybackslash}X}

\centering{(a) Semantic labels and their relative sizes on average (\%).}
\begin{tabularx}{\linewidth}{RllRllRllRll}
\toprule
1. & Cerebral grey & (50.24) & 2. & 3rd ventricle & (0.09) & 3. & 4th ventricle & (0.15) & 4. & Brainstem & (1.46) \\
\midrule
5. & CVL I-V & (0.39) & 6. & CVL VI-VII & (0.19) & 7. & CVL VIII-X & (0.26) & 8. & Accumbens & (0.07) \\
\midrule
9. & Amygdala & (0.21) & 10. & Caudate & (0.54) & 11. & Cerebellar grey & (8.19) & 12. & Cerebellar white & (2.06) \\
\midrule
13. & Cerebral white & (31.23) & 14. & Hippocampus & (0.58) & 15. & Inf. lateral vent. & (0.09) & 16. & Lateral ventricle & (2.11) \\
\midrule
17. & Pallidum & (0.25) & 18. & Putamen & (0.73) & 19. & Thalamus & (1.19) \\
\bottomrule
\medskip
\end{tabularx}

\centering{(b) Average Dice coefficients (mean$\pm$std\%) with respective to the ground truth.}
\begin{tabularx}{\linewidth}{lRlRlRlRlRlRlRl}
\toprule
\multicolumn{15}{c}{Proposed network with linear Dice loss, logarithmic Dice loss, and weighted cross-entropy} \\
\midrule
\multirow{3}{0.18\linewidth}{$\mathbf{E}\left[1 - \mathrm{Dice}_i\right]$ (\ref{eq:dice_loss})}
& 1. & 87$\pm$1 & 2. & 47$\pm$38 & 3. & 32$\pm$40 & 4. & 72$\pm$36 & 5. & 50$\pm$41 & 6. & 30$\pm$37 & 7. & 31$\pm$38 \\
& 8. & 0$\pm$0 & 9. & 0$\pm$0 & 10. & 34$\pm$42 & 11. & 88$\pm$1 & 12. & 86$\pm$1 & 13. & 88$\pm$1 & 14. & 32$\pm$39 \\
& 15. & 0$\pm$0 & 16. & 54$\pm$44 & 17. & 0$\pm$0 & 18. & 51$\pm$42 & 19. & 35$\pm$43 & \multicolumn{4}{l}{\ \textbf{Average: 43$\pm$11}} \\
\midrule
\multirow{3}{0.18\linewidth}{$L_\mathrm{Dice}(\gamma = 1)$ (\ref{eq:dice_loss})}
& 1. & 84$\pm$1 & 2. & 61$\pm$30 & 3. & 83$\pm$2 & 4. & 90$\pm$1 & 5. & 81$\pm$2 & 6. & 73$\pm$2 & 7. & {\color{blue}\textbf{78$\pm$2}} \\
& 8. & 68$\pm$2 & 9. & 74$\pm$2 & 10. & 85$\pm$1 & 11. & 87$\pm$1 & 12. & 85$\pm$1 & 13. & 88$\pm$1 & 14. & 79$\pm$2 \\
& 15. & 59$\pm$3 & 16. & 89$\pm$1 & 17. & 79$\pm$1 & 18. & 86$\pm$2 & 19. & 88$\pm$1 & \multicolumn{4}{l}{\ \textbf{Average: 80$\pm$2}} \\
\midrule
\multirow{3}{0.18\linewidth}{$L_\mathrm{Cross}(\gamma = 1)$ (\ref{eq:entropy_loss})}
& 1. & 87$\pm$1 & 2. & 56$\pm$5 & 3. & 79$\pm$3 & 4. & 86$\pm$2 & 5. & 76$\pm$3 & 6. & 67$\pm$2 & 7. & 73$\pm$6 \\
& 8. & 59$\pm$4 & 9. & 65$\pm$4 & 10. & 83$\pm$2 & 11. & 87$\pm$2 & 12. & 85$\pm$1 & 13. & 89$\pm$1 & 14. & 75$\pm$3 \\
& 15. & 54$\pm$6 & 16. & 89$\pm$1 & 17. & 76$\pm$3 & 18. & 84$\pm$1 & 19. & 86$\pm$1 & \multicolumn{4}{l}{\ \textbf{Average: 77$\pm$2}} \\
\midrule
\multicolumn{15}{c}{Proposed network with $L_\mathrm{Exp}$ at different values of $\gamma$} \\
\midrule
\multirow{3}{0.18\linewidth}{$L_\mathrm{Exp}(\gamma = 1)$ (\ref{eq:weighted_loss})}
& 1. & 87$\pm$2 & 2. & {\color{blue}\textbf{78$\pm$3}} & 3. & {\color{blue}\textbf{84$\pm$1}} & 4. & 90$\pm$1 & 5. & {\color{blue}\textbf{82$\pm$1}} & 6. & 74$\pm$2 & 7. & 78$\pm$3 \\
& 8. & 68$\pm$3 & 9. & {\color{blue}\textbf{75$\pm$1}} & 10. & 83$\pm$3 & 11. & 87$\pm$1 & 12. & {\color{blue}\textbf{86$\pm$0}} & 13. & 89$\pm$1 & 14. & 80$\pm$1 \\
& 15. & {\color{blue}\textbf{64$\pm$1}} & 16. & 90$\pm$1 & 17. & 80$\pm$2 & 18. & 86$\pm$2 & 19. & 88$\pm$1 & \multicolumn{4}{l}{\ \textbf{Average: 81$\pm$1}} \\
\midrule
\multirow{3}{0.18\linewidth}{$L_\mathrm{Exp}(\gamma = 2)$ (\ref{eq:weighted_loss})} & 1. & 79$\pm$7 & 2. & 61$\pm$15 & 3. & 74$\pm$6 & 4. & 75$\pm$10 & 5. & 67$\pm$12 & 6. & 62$\pm$8 & 7. & 66$\pm$10 \\
& 8. & 52$\pm$17 & 9. & 56$\pm$15 & 10. & 64$\pm$12 & 11. & 78$\pm$8 & 12. & 78$\pm$7 & 13. & 84$\pm$4 & 14. & 64$\pm$11 \\
& 15. & 46$\pm$10 & 16. & 77$\pm$10 & 17. & 60$\pm$16 & 18. & 67$\pm$15 & 19. & 67$\pm$15 & \multicolumn{4}{l}{\ \textbf{Average: 67$\pm$11}} \\
\midrule
\multirow{3}{0.18\linewidth}{$L_\mathrm{Exp}(\gamma = 0.3)$ (\ref{eq:weighted_loss})}
& 1. & {\color{blue}\textbf{88$\pm$1}} & 2. & 77$\pm$2 & 3. & {\color{blue}\textbf{84$\pm$1}} & 4. & {\color{blue}\textbf{91$\pm$1}} & 5. & {\color{blue}\textbf{82$\pm$1}} & 6. & {\color{blue}\textbf{74$\pm$1}} & 7. & {\color{blue}\textbf{78$\pm$2}} \\
& 8. & {\color{blue}\textbf{69$\pm$2}} & 9. & 75$\pm$2 & 10. & {\color{blue}\textbf{86$\pm$1}} & 11. & {\color{blue}\textbf{89$\pm$1}} & 12. & 86$\pm$1 & 13. & {\color{blue}\textbf{89$\pm$0}} & 14. & {\color{blue}\textbf{81$\pm$1}} \\
& 15. & 62$\pm$5 & 16. & {\color{blue}\textbf{91$\pm$1}} & 17. & {\color{blue}\textbf{80$\pm$1}} & 18. & {\color{blue}\textbf{87$\pm$1}} & 19. & {\color{blue}\textbf{89$\pm$1}} & \multicolumn{4}{l}{\ {\color{blue}\textbf{Average: 82$\pm$1}}} \\
\midrule
\multicolumn{15}{c}{V-Net with the best $L_\mathrm{Exp}$ at $\gamma = 0.3$} \\
\midrule
\multirow{3}{0.18\linewidth}{V-Net\\ $L_\mathrm{Exp}(\gamma = 0.3)$   (\ref{eq:weighted_loss})}
& 1. & 84$\pm$2 & 2. & 67$\pm$7 & 3. & 80$\pm$4 & 4. & 87$\pm$4 & 5. & 78$\pm$3 & 6. & 67$\pm$5 & 7. & 73$\pm$6 \\
& 8. & 59$\pm$7 & 9. & 65$\pm$5 & 10. & 72$\pm$5 & 11. & 85$\pm$2 & 12. & 82$\pm$4 & 13. & 86$\pm$2 & 14. & 72$\pm$7 \\
& 15. & 48$\pm$8 & 16. & 82$\pm$6 & 17. & 70$\pm$7 & 18. & 75$\pm$6 & 19. & 78$\pm$6 & \multicolumn{4}{l}{\ \textbf{Average: 74$\pm$4}} \\
\bottomrule
\end{tabularx}
\end{table*}


\subsection{Results and Discussion}

Table \ref{table:brain}(b) shows the Dice coefficients averaged from the five experiments. The linear Dice loss ($\mathbf{E}\left[1 - \mathrm{Dice}_i\right]$) had the worst performance. It performed well with the relatively large structures such as the gray and white matters, but the performance decreased with the sizes of the structures. The very small structures, such as the nucleus accumbens and amygdala, were missed in all experiments. In contrast, the logarithmic Dice loss ($L_\mathrm{Dice}(\gamma = 1)$) provided much better results, though the large standard deviation of label 2 indicates that there were misses. We also performed experiments with the weighted cross-entropy ($L_\mathrm{Cross}(\gamma = 1)$), whose performance was better than the linear Dice loss but worse than the logarithmic Dice loss. The weighted sum of the logarithmic Dice loss and weighted cross-entropy ($L_\mathrm{Exp}(\gamma = 1)$) outperformed the individual losses, and it provided the second best results among the tested cases. As discussed in Section \ref{sec:loss}, $L_\mathrm{Exp}(\gamma = 2)$ was ineffective even on larger structures. This is consistent with our observation in Fig. \ref{fig:plot_loss} that the loss function is over suppressed when the accuracy is getting higher. In contrast, $L_\mathrm{Exp}(\gamma = 0.3)$ gave the best results. Although it only performed slightly better than $L_\mathrm{Exp}(\gamma = 1)$ in terms of the means, the smaller standard deviations indicate that it was also more precise.

\begin{figure}[t]
    \footnotesize
    \centering
    \begin{minipage}[b]{0.6\linewidth}
      \centering
      \includegraphics[width=1\linewidth]{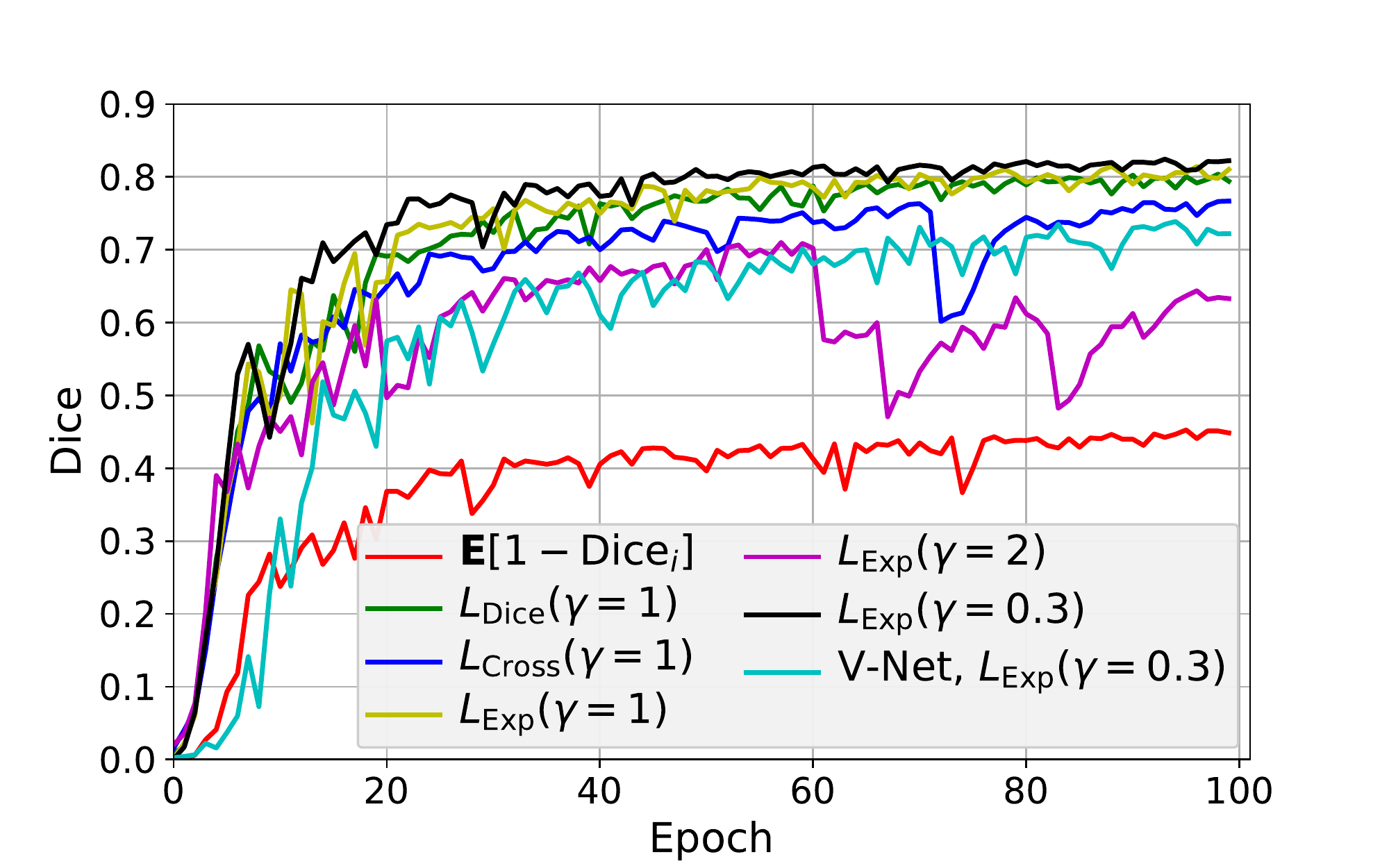} \\
    \end{minipage}
    \caption{Validation Dice coefficients vs. epoch, averaged from five experiments.}
    \label{fig:plot_dice}
\end{figure}

\begin{figure}[t]
    \fontsize{5}{6}\selectfont
    \centering
    \begin{minipage}[t]{0.115\linewidth}
      \centering
      \includegraphics[width=1\linewidth]{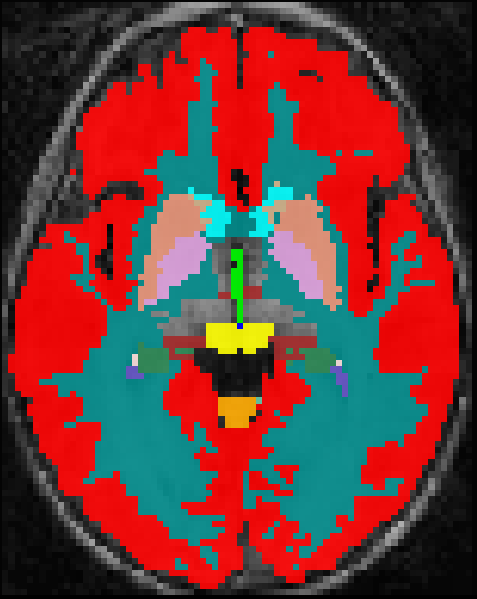} \\
      \includegraphics[width=1\linewidth]{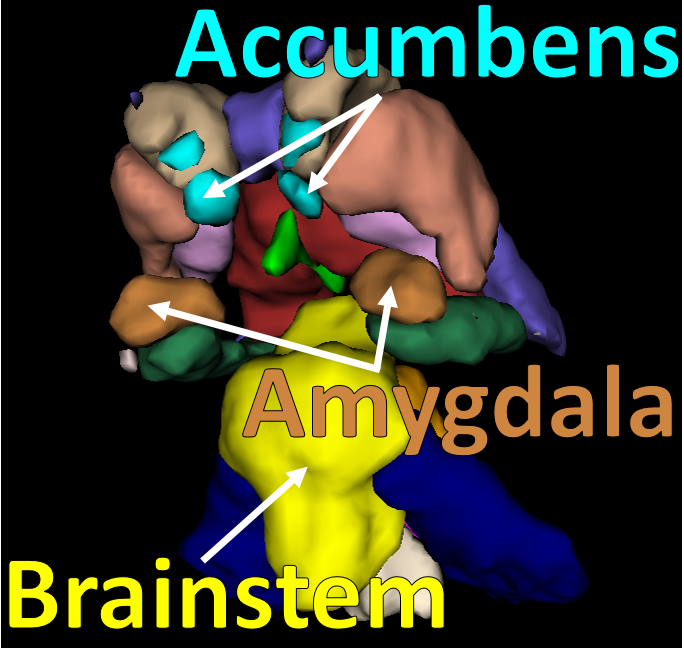} \\
      \centering{Ground truth \\ \phantom{a}}
    \end{minipage}
    \vrule\
    \begin{minipage}[t]{0.115\linewidth}
      \centering
      \includegraphics[width=1\linewidth]{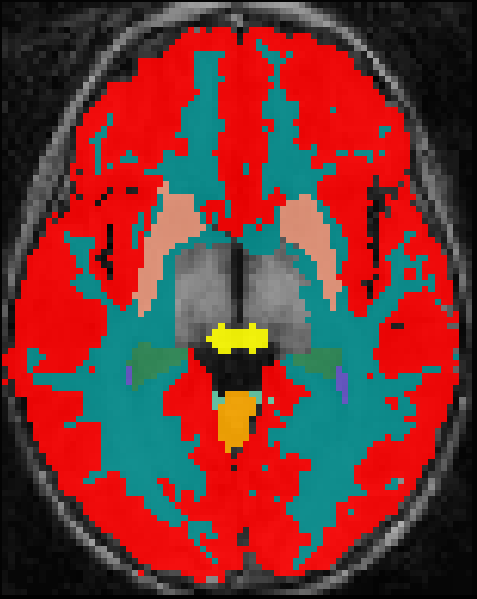} \\
      \includegraphics[width=1\linewidth]{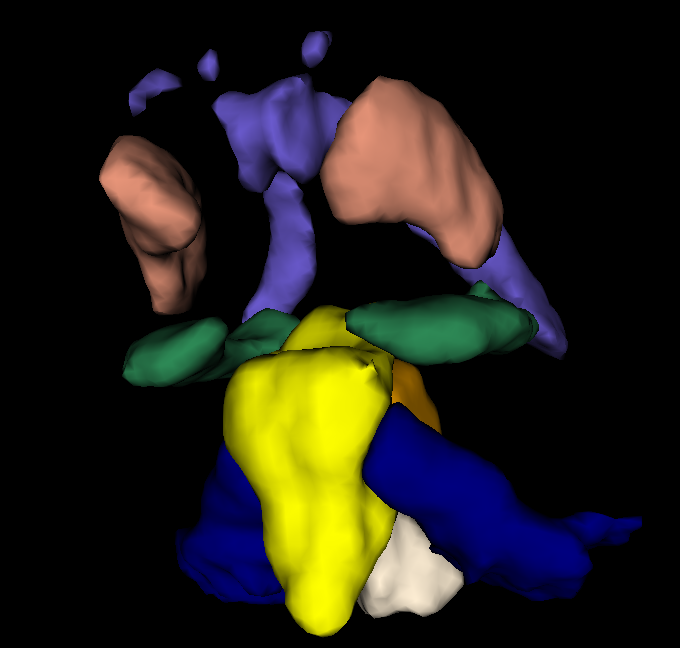} \\
      \centering{$\mathbf{E}\left[1 - \mathrm{Dice}_i\right]$ \\ \phantom{a} Dice = 51\%}
    \end{minipage}
    \begin{minipage}[t]{0.115\linewidth}
      \centering
      \includegraphics[width=1\linewidth]{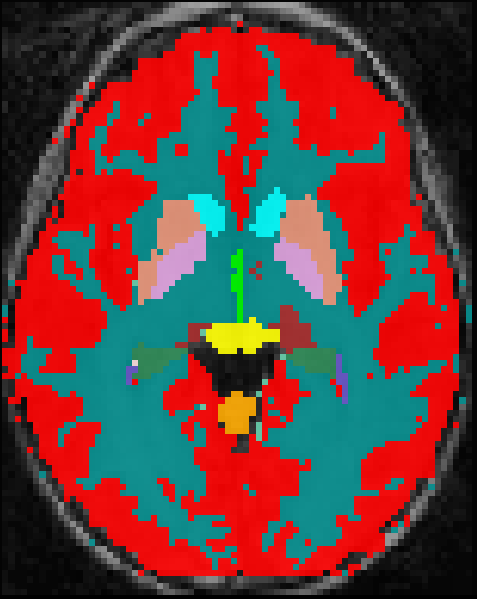} \\
      \includegraphics[width=1\linewidth]{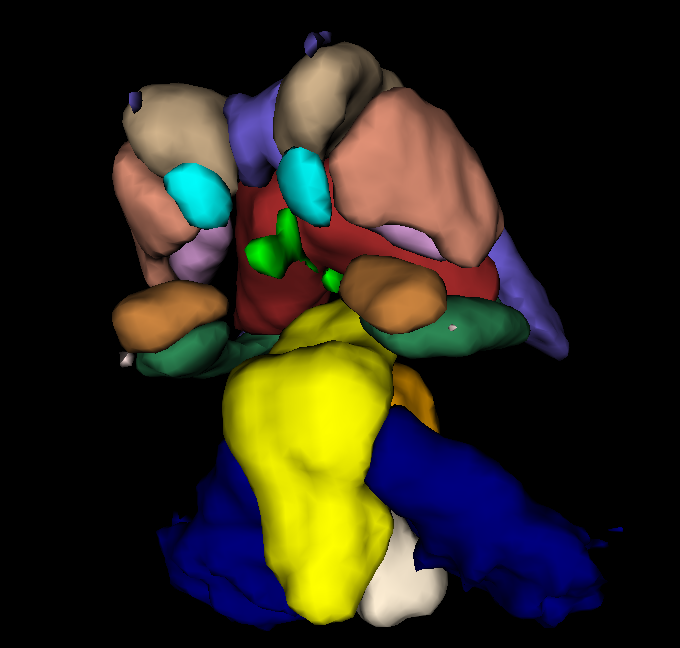} \\
      \centering{$L_\mathrm{Dice}$\\ $(\gamma = 1)$ \\ Dice = 81\%}
    \end{minipage}
    \begin{minipage}[t]{0.115\linewidth}
      \centering
      \includegraphics[width=1\linewidth]{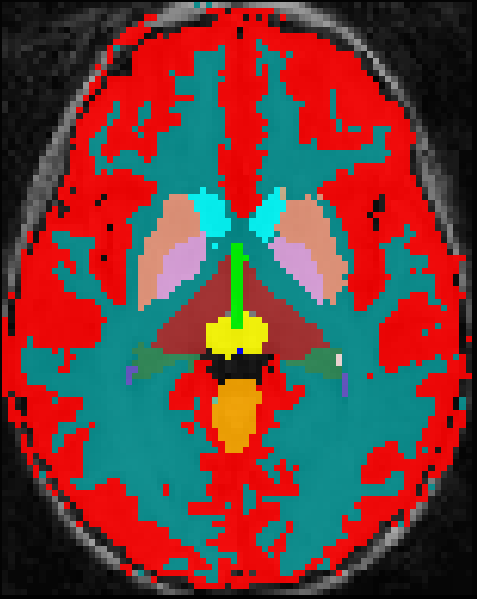} \\
      \includegraphics[width=1\linewidth]{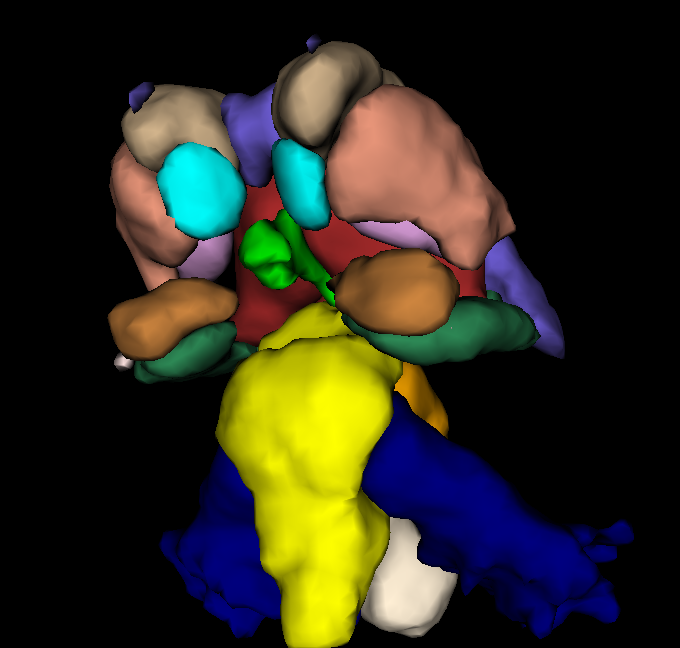} \\
      \centering{$L_\mathrm{Cross}$\\ $(\gamma = 1)$ \\ Dice = 76\%}
    \end{minipage}
    \vrule\
    \smallskip
    \begin{minipage}[t]{0.115\linewidth}
      \centering
      \includegraphics[width=1\linewidth]{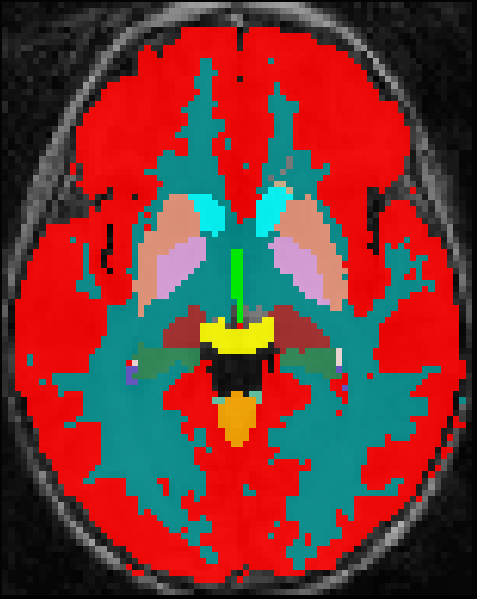} \\
      \includegraphics[width=1\linewidth]{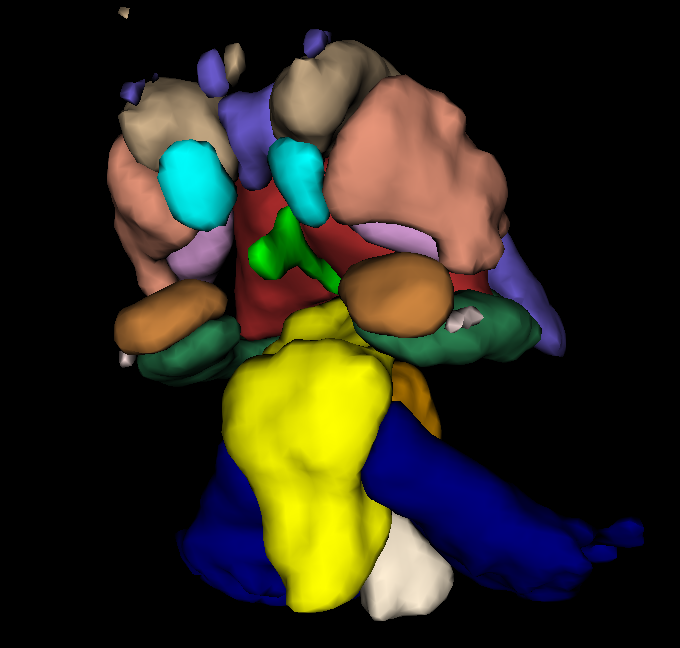} \\
      \centering{$L_\mathrm{Exp}$\\ $(\gamma = 1)$ \\ Dice = 80\%}
    \end{minipage}
    \begin{minipage}[t]{0.115\linewidth}
      \centering
      \includegraphics[width=1\linewidth]{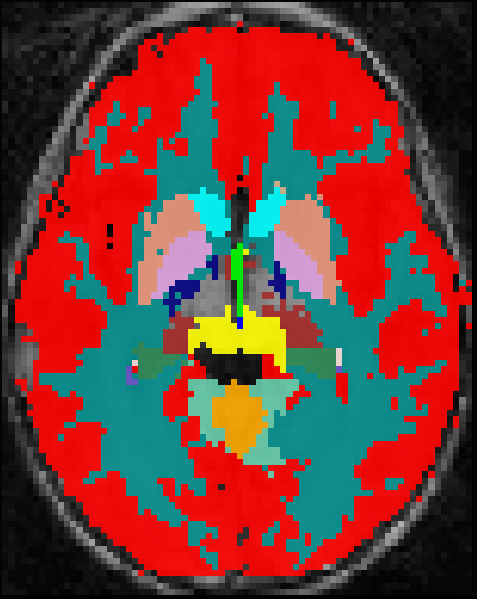} \\
      \includegraphics[width=1\linewidth]{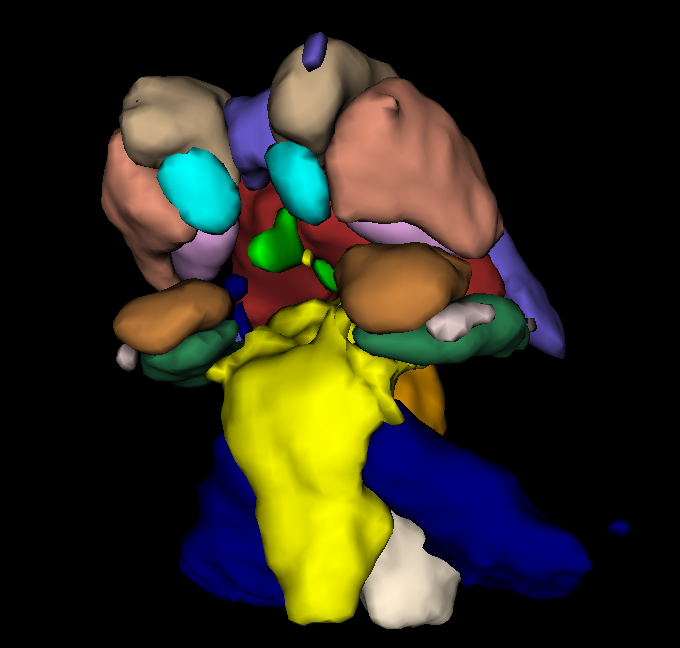} \\
      \centering{$L_\mathrm{Exp}$\\ $(\gamma = 2)$ \\ Dice = 76\%}
    \end{minipage}
    \begin{minipage}[t]{0.115\linewidth}
      \centering
      \includegraphics[width=1\linewidth]{{{A_expDiceEntropy0.3}}} \\
      \includegraphics[width=1\linewidth]{{{3D_expDiceEntropy0.3}}} \\
      \centering{$L_\mathrm{Exp}$\\ $(\gamma = 0.3)$ \\ Dice = 81\%}
    \end{minipage}
    \vrule\
    \begin{minipage}[t]{0.115\linewidth}
      \centering
      \includegraphics[width=1\linewidth]{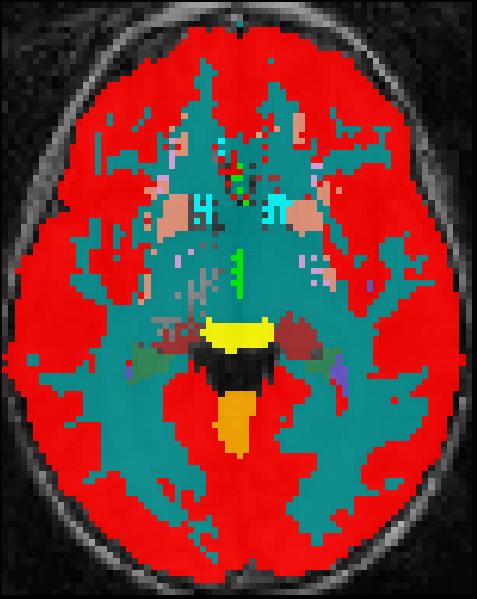} \\
      \includegraphics[width=1\linewidth]{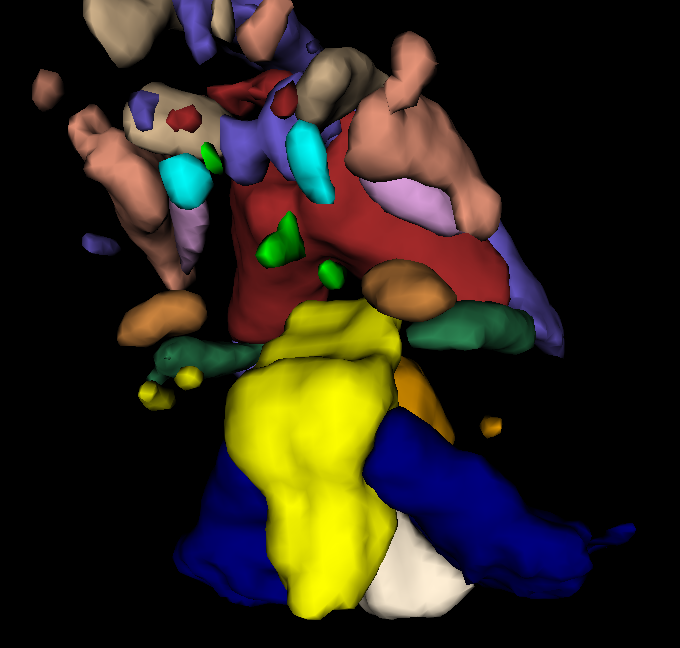} \\
      \centering{V-Net \\ $(\gamma = 0.3)$ \\ Dice = 65\%}
    \end{minipage}
    \caption{Visualization of an example. Top: axial view. Bottom: 3D view with the cerebral grey, cerebral white, and cerebellar grey matters hidden for better illustration.}
    \label{fig:visualization}
\end{figure}

When applying the best loss function to the V-Net, its performance was only better than the linear Dice loss and $L_\mathrm{Exp}(\gamma = 2)$. This shows that our proposed network architecture performed better than the V-Net on this problem.

Fig. \ref{fig:plot_dice} shows the validation Dice coefficients vs. epoch, averaged from the five experiments. Instead of the losses, we show the Dice coefficients as their magnitudes were consistent among cases. Similar to Table \ref{table:brain}(b), the logarithmic Dice loss, $L_\mathrm{Exp}(\gamma = 1)$, and $L_\mathrm{Exp}(\gamma = 0.3)$ had good convergence and performance, with $L_\mathrm{Exp}(\gamma = 0.3)$ performed slightly better. These three cases converged at about 80 epochs. The weighted cross-entropy and $L_\mathrm{Exp}(\gamma = 2)$ were more fluctuating. The linear Dice loss also converged at about 80 epochs but with a much smaller Dice coefficient. Comparing between the V-Net and the proposed network with $L_\mathrm{Exp}(\gamma = 0.3)$, the V-Net had worse convergence especially at the earlier epochs. This shows that the proposed network had better convergence.

Fig. \ref{fig:visualization} shows the visualization of an example. There are two obvious observations. First of all, consistent with Table \ref{table:brain}(b), the linear Dice loss missed some small structures such as the nucleus accumbens and amygdala, though it performed well on large structures. Secondly, the segmentation of the V-Net deviated a lot from the ground truth. The logarithmic Dice loss, $L_\mathrm{Exp}(\gamma = 1)$, and $L_\mathrm{Exp}(\gamma = 0.3)$ had the best segmentations and average Dice coefficients. The weighted cross-entropy had the same average Dice coefficient as $L_\mathrm{Exp}(\gamma = 2)$, though the weighted cross-entropy over-segmented some structures such as the brainstem, and $L_\mathrm{Exp}(\gamma = 2)$ had a noisier segmentation.

Comparing the efficiencies between the proposed network and the V-Net, the proposed network had around 5 million parameters while the V-Net had around 71 million parameters, a 14-fold difference. Furthermore, the proposed network only took about 0.4 s on average to segment a 128$\times$128$\times$128 volume, while the V-Net took about 0.9 s. Therefore, the proposed network was more efficient.





\section{Conclusion}

In this paper, we propose a network architecture optimized for 3D image segmentation, and a loss function for segmenting very small structures. The proposed network architecture has only about 1/14 of the parameters of, and is twice as fast as, the V-Net. For the loss function, the logarithmic Dice loss outperforms the linear Dice loss, and the weighted sum of the logarithmic Dice loss and the weighted cross-entropy outperforms the individual losses. With the introduction of the exponential form, the nonlinearities of the loss functions can be further controlled to improve the accuracy and precision of segmentation.

\bibliographystyle{splncs03}
\bibliography{Ref}

\end{document}